%% file: main_arxiv.tex
\documentclass[letterpaper]{article} 
\usepackage{aaai24}  
\usepackage{times}  
\usepackage{helvet}  
\usepackage{courier}  
\usepackage[hyphens]{url}  
\usepackage{graphicx} 
\usepackage{natbib}  
\usepackage{caption}
\usepackage{bibentry}
\usepackage{algorithmic}
\usepackage{algorithm}
\usepackage{newfloat}
\usepackage{multirow}
\usepackage{listings}
\usepackage{booktabs}
\usepackage{tabularx}
\usepackage{amsmath}
\usepackage{amsfonts,amssymb} 
\usepackage{xcolor}         

\floatstyle{ruled}
\newfloat{listing}{tb}{lst}{}
\floatname{listing}{Listing}
\DeclareCaptionStyle{ruled}{labelfont=normalfont,labelsep=colon,strut=off} 
\usepackage{graphicx}
\usepackage{amssymb}
\usepackage{changepage}
\usepackage{bm}
\usepackage{tabularray}
\title{Semantic Equivariant Mixup}
\author{
    Zongbo Han\textsuperscript{\rm 1*}, Tianchi Xie\textsuperscript{\rm 1*}\thanks{Equal contribution. }, Bingzhe Wu\textsuperscript{\rm 2}, Qinghua Hu\textsuperscript{\rm 1}, Changqing Zhang\textsuperscript{\rm 1}
}
\affiliations{
    \textsuperscript{\rm 1} College of Intelligence and Computing, Tianjin University\\
    \textsuperscript{\rm 2} Tencent AI Lab\\ 
}

\begin{document}
\maketitle
\begin{abstract}
Mixup is a well-established data augmentation technique, which can extend the training distribution and regularize the neural networks by creating ``mixed'' samples based on the \textit{label-equivariance} assumption, i.e., a proportional mixup of the input data results in the corresponding labels being mixed in the same proportion. However, previous mixup variants may fail to exploit the label-independent information in mixed samples during training, which usually contains richer semantic information. To further release the power of mixup, we first improve the previous label-equivariance assumption by the semantic-equivariance assumption, which states that the proportional mixup of the input data should lead to the corresponding representation being mixed in the same proportion. Then a generic mixup regularization at the representation level is proposed, which can further regularize the model with the semantic information in mixed samples. At a high level, the proposed semantic equivariant mixup (\textsc{sem}) encourages the structure of the input data to be preserved in the representation space, i.e., the change of input will result in the obtained representation information changing in the same way. Different from previous mixup variants, which tend to over-focus on the label-related information, the proposed method aims to preserve richer semantic information in the input with semantic-equivariance assumption, thereby improving the robustness of the model against distribution shifts. We conduct extensive empirical studies and qualitative analyzes to demonstrate the effectiveness of our proposed method. The code of the manuscript is in the supplement. 
\end{abstract}

\section{Introduction}
\input{1_introduction}

\section{Related works}
\input{2_relatedworks}

\section{Method}
\input{3_method}

\section{Experiments}
\input{4_experiments}

\section{Conclusion and Limitation}
In this paper, we focus on improving classical mixup algorithms by exploiting semantic information in mixed samples. We first summarize the label equivariance assumption shared by mixup variants and establish its relation to equivariance in deep learning. Then we propose the semantic equivariance assumption, a generalization of label equivariance at the semantic level. Based on this assumption, we can exploit the comprehensive semantic information in mixed samples by regularizing the representation of mixed samples. We propose a simple but effective constraint term, which is a general mixup improvement and thus can be used with previous mixup methods. We conduct extensive experiments to illustrate the superiority of the proposed method in terms of classification performance, generalization, and out-of-distribution detection. Meanwhile, we conduct a qualitative analysis to verify the effect of the the proposed \textsc{sem}. At the same time, we acknowledge that, as shown in Alg.~\ref{alg:semix}, SEM requires more computational overhead in the training phase to obtain $r_i$ and $r_j$, but the same overhead in the inference phase as other mixup algorithms. In the future, we will continue to explore utilizing semantic information in mixed samples while reducing computation cost to improve robustness and generalization. 

\bibliography{main}


\end{document}

%% file: 1_introduction.tex
\begin{figure}[ht]
    \centering
    \includegraphics[width=0.5\textwidth]{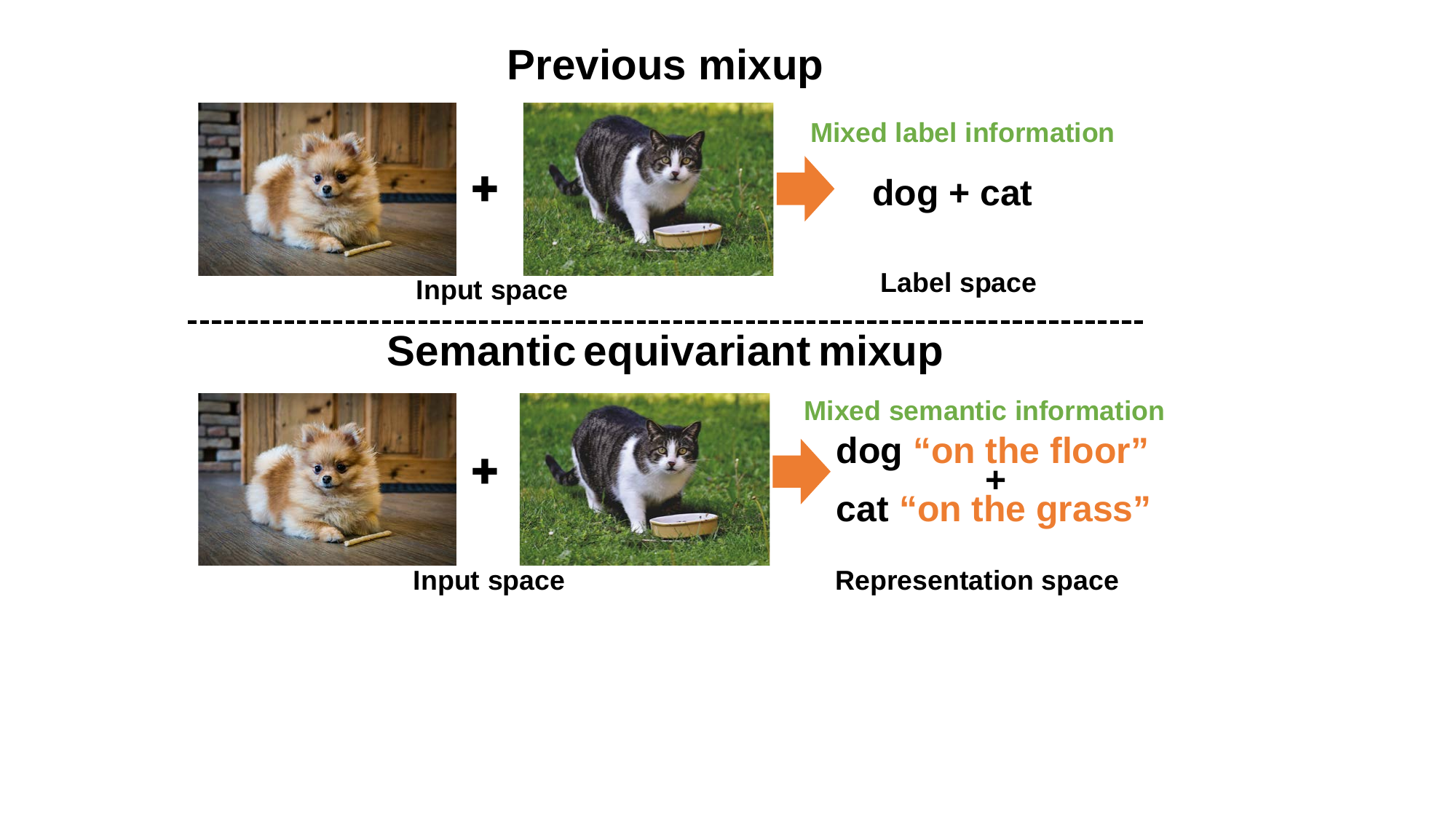}
    \caption{Motivation of the proposed method. The previous mixup only focused on the label information in the mixed samples, while the proposed method aim to leverage semantic information in the representation as the nuanced and comprehensive supervision to improve the previous mixup.}
    \label{fig:motivation}
\end{figure}
Modern deep neural networks face the challenges of overfitting or even memorizing the training samples due to their powerful learning capacity, resulting in superior performance on the training set but poor generalization on test data \cite{goodfellow2016deep}. To address this problem, data augmentation is widely employed during deep neural network training, which constructs more training samples by perturbing the samples in the training set, such as rotating and cropping images \cite{shorten2019survey}. Beyond these naive augmentation strategies, mixup \cite{zhang2018mixup} proposes a more general approach to create more diverse training samples by linearly interpolating input data and corresponding labels, thus extending the training distribution in a relatively reasonable way and preventing the deep network from overfitting the training data. 

Due to the simplicity and effectiveness, mixup-based methods have gained popularity in various data types and tasks \cite{yun2019cutmix,kim2020puzzle,kim2023exploring,sahoo2021contrast,wang2021mixup,han2022g,verma2019manifold,han2022umix,han2023reweighted,zhou2020domain,mroueh2021fair,zhang2022and,thulasidasan2019mixup}. 
Existing mixup variants \cite{zhang2018mixup,yun2019cutmix} typically share the same assumption called label-equivariance assumption, which states that \textit{a proportional mixup of the input data results in the corresponding labels being mixed in the same proportion}. For example, vanilla mixup \cite{zhang2018mixup} assumes that conducting linear interpolation $\lambda{x}_1+(1-\lambda){x}_2$ between a pair of training points ${({x}_1,y_1), ({x}_2,y_2)}$ in the input space with a random interpolation ratio $\lambda$ will lead to interpolation on the label space in a same ratio, i.e, $\lambda y_1+(1-\lambda)y_2$. Mixup leverages this assumption to create mixed samples and regularize the deep neural networks. Previous mixup variants mainly focus on satisfying the label-equivariance assumption by carefully combining input data without exploiting other semantic information in mixed samples. For example, in mixup for image data, previous methods mainly focus on cropping label-related regions from one image and pasting them onto label-independent regions of another image \cite{kim2020puzzle,beckham2019adversarial,walawalkar2020attentive}. To achieve this goal, additional modules such as saliency detection \cite{kim2020puzzle},  attention mechanism \cite{walawalkar2020attentive} and adversarial network \cite{beckham2019adversarial} are introduced.

However, previous mixup variants overlook the label-independent information in the mixed samples, which typically contains richer semantic information. For example, as shown in Fig.~\ref{fig:motivation}, the mixed image contains not only label-related information, but also other semantic information. Meanwhile, previous studies have revealed that over-emphasizing label-related information may hurt model training. Specifically, the simplicity bias can be introduced in the model, causing it to rely excessively on simpler features and neglect other semantic information \cite{shah2020pitfalls,pezeshki2021gradient}. Besides, this can lead to the neural network taking shortcuts instead of learning the intended solution to identify label-related information, thereby reducing the generalization of the model \cite{geirhos2020shortcut}. To this end, the proposed method (\textsc{sem}) aims to generalize the label-equivariance assumption to the semantic level and provide nuanced and
comprehensive supervision, enabling better regularization of the model using the richer semantic information in the mixed samples. 


Inspired by the recent advancements in equivariant neural networks \cite{cohen2016group,dangovski2021equivariant,dangovski2022equivariant,satorras2021n}, we first establish the connection between the equivariance in deep learning and the mixup. Then we introduce the generalized semantic-equivariance assumption in the representation space that states 
\begin{adjustwidth}{1em}{1em}
\emph{a proportional mixup of the input results in a combination of the representations in the same proportion.}
\end{adjustwidth}
Building upon this assumption, we can improve previous mixup methods by introducing a simple yet effective representation regularization that encourage changes in the original input space to also occur in the representation space. In contrast to previous mixup-based methods, which only considers label information in the mixed samples, the proposed method focuses on the semantic equivariance in the representation space, providing a stronger constraint that preserves richer information and mitigates the simplicity bias and the shorcut problems of neural networks. 
Our contributions can be summarized as follows:
\begin{itemize}
    \item To the best of our knowledge, we are the first to summarize the label-equivariance assumption underlying mixup-based methods and demonstrate the connection between the mixup and equivariance in deep learning. 
    \item We propose a semantic-equivariance assumption that improves the mixup's label equivariance assumption. Based on the semantic-equivariance assumption, we introduce a simple and effective regularization term that can cooperate with most previous mixup variants and encourages the model to consider semantic information in the input. 
    \item Extensive experiments show that our proposed method can cooperate well with previous mixup algorithms and achieve better performance.
\end{itemize}

%% file: 2_relatedworks.tex
\textbf{Mixup-based methods.} Mixup \cite{zhang2018mixup} is a widely used data augmentation technique that can improve the robustness and generalization of deep neural networks by creating mixed samples with interpolation between different training samples \cite{yun2019cutmix,chidambaram2022provably,park2022unified}. Going beyond the empirical risk minimization paradigm, mixup can be regarded as a special case of vicinal risk minimization \cite{chapelle2000vicinal} that ensures the model's good performance  not only on the training set but also in the vicinal space of the training set. Most of the previous mixup variants focus on designing how to mix different samples so that the mixed samples are helpful for neural network training \cite{yun2019cutmix, kim2020puzzle, beckham2019adversarial, verma2019manifold, wang2021mixup, han2022g, yao2022improving}. Puzzlemix \cite{kim2020puzzle} selects areas of sufficient saliency information from each image and combines them to better preserve the static information of both sides. SaliencyMix \cite{uddin2020saliencymix} employs a saliency map to select informative image patches and avoid introducing noise. Alignmix \cite{venkataramanan2021alignmixup} performs mixing in the feature space by geometrically aligning two features before combining them to create synthetic data. In addition, mixup variants have been shown to be effective on a variety of tasks, including fairness machine learning \cite{han2022umix,han2023reweighted,mroueh2021fair}, domain generalization \cite{zhou2020domain,yao2022improving}, confidence calibration \cite{zhang2022and,thulasidasan2019mixup}. 

\textbf{Comparison with existing works.}  While the label-equivariance assumption is a key component in the construction of mixed samples, it has received relatively little attention in existing studies. In contrast, we aim to build on this assumption by incorporating the semantic information of mixed samples to further enhance the robustness and generalization of the model, thus improving the robustness and generalization of the model. Among these methods, manifold mixup \cite{verma2019manifold} is the most related work to ours. Specifically, manifold mixup creates mixed samples by interpolating in the latent space to obtain better representations, which may limit its application, such as its inability to perform mixing in the image space. In contrast, \textsc{sem} focuses on the core hypothesis of mixup and proposes an intuitive semantic equivariance assumption that proportional mixup of inputs should lead to proportional mixup of representations, thereby exploiting the comprehensive semantic information in the mixed samples. Besides, in practice, \textsc{sem} also can cooperate with most input space mixing approaches like vanilla mixup, CutMix, etc. 

\textbf{Equivariance in deep learning.}  
Equivariance is a critical property that can be exploited to improve the structure of deep neural network representations, which has been applied in classic models such as alexnet \cite{lenc2015understanding,krizhevsky2017imagenet}, capsule network \cite{hinton2011transforming} and AlphaFold \cite{jumper2021highly}. The fundamental concept is that changes to the neural network input, such as translation and rotation, will be reflected in the representation. Prior researches have primarily focused on enhancing the neural network equivariance by designing a strict equivariant neural networks or using regularization techniques to promote their equivariance.
Specifically, existing methods \cite{cohen2016group,pmlr-v119-romero20a,marcos2017rotation,cohen2018spherical,thomas2018tensor,cohen2019general,keriven2019universal,cohen2019gauge,fuchs2020se} employ equivariance to improve the original neural network to preserve more structure of the input. 
Moreover, recent works have enhanced the representation learning of models by introducing equivariance in contrastive learning \cite{devillers2023equimod,dangovski2022equivariant}. In this paper, we analyze the connection between mixup and isovariance and propose an improvement to previous mixup methods by leveraging the semantic information of mixed samples to enhance the label-equivariance assumption.

%% file: 3_method.tex
In this section, we introduce the proposed Semantic Equivariant Mixup (\textsc{sem}), which leverages label-independent semantic information in mixed samples to provide more nuanced and comprehensive supervision information for deep learning networks, thereby mitigating simplicity bias and shortcut learning resulting from over-emphasizing label information. Specifically, after presenting necessary preliminary concepts, we demonstrate that mixup and equivariant deep neural networks share similar motivations. We then detail how to incorporate semantic information into mixed samples to improve previous mixup methods.

\subsection{Preliminary}
\label{sec:preliminary}
\textbf{Setup.} Let $\mathcal{X}$ and $\mathcal{Y}$ denote the input and label space, respectively, and assume that $f^\star:\mathcal{X}\rightarrow \mathcal{Y}$ is an ideal labeling  function. Given a training dataset $\mathcal{D}=\{x_i,y_i\}_{i=1}^N$ with $N$ samples drawn from the training distribution, where $x\in\mathcal{X}$ represents the input data and $y=f^\star(x)\in\mathcal{Y}$ is the label presented as a one-hot vector in the classification task or a scalar in the regression task, our goal is to train a model $f_\theta:\mathcal{X}\rightarrow \mathcal{Y}$ parameterized by $\theta\in\Theta$ that performs well on the test set. To accomplish this, the well-known empirical risk minimization (ERM) usually minimizes the following expected risk defined as
\begin{equation}
\mathbb{E}_\mathcal{D} [\ell(f_{\theta}(x_i),y_i)],
\end{equation}
where $\ell:\Theta\times\mathcal{X}\times\mathcal{Y}\rightarrow\mathbb{R}$ denotes the loss function to calculate the distance between $f_{\theta}(x)$ and $y$, and the cross-entropy loss is usually employed in classification and the mean square error loss is employed in regression task.

\textbf{Previous mixup.} Although deep learning paradigms based on empirical risk minimization achieve superior performance, deep neural networks often suffer from overfitting the training dataset due to their strong learning capacity. To this end, mixup-based methods create mixed samples by combining pairs of training examples to extend the training distribution and prevent the deep neural network from overfitting the training data \cite{zhang2018mixup,yun2019cutmix, kim2020puzzle, beckham2019adversarial, verma2019manifold, wang2021mixup, han2022g, yao2022improving}. Typically, vanilla mixup \cite{zhang2018mixup} creates mixed samples using linear interpolation between different samples as follows,
\begin{equation}
\label{eq:mixup}
\widetilde{x}_{i,j}=\lambda x_i+ (1-\lambda) x_j,
\widetilde{y}_{i,j}=\lambda y_i+(1-\lambda) y_j,
\end{equation}
where $\lambda\in[0,1]$ is sampled from a Beta distribution $Beta(\alpha,\alpha)$ with hyperparameter $\alpha$. To reduce the noise introduced during image interpolation, CutMix \cite{yun2019cutmix} randomly selects regions from an image and pastes them onto another image to create training samples as follows,
\begin{equation}
\label{eq:cutmix}
\widetilde{x}_{i,j}=B_\lambda \odot x_i+(1-B_\lambda)\odot x_j,
\widetilde{y}_{i,j}=\lambda y_i+(1-\lambda) y_j,
\end{equation}
where $B_\lambda$ is a random binary mask indicating the location of the cropped region, $\lambda$ represents the cropped area ratio and is also sampled from a $Beta(\alpha,\alpha)$ distribution, and $\odot$ denotes the element-wise product. Overall, previous mixup variants mainly focus on improving the mixing process to extend the training distribution \cite{yun2019cutmix,kim2020puzzle,kim2023exploring,sahoo2021contrast,verma2019manifold,wang2021mixup,han2022g}.

%
%

\textbf{Equivariance in deep learning.} Equivariance is an important inductive bias in deep learning and is presented in many classic methods such as AlexNet \cite{lenc2015understanding,krizhevsky2017imagenet}, capsule network \cite{hinton2011transforming}, and AlphaFold \cite{jumper2021highly}.
Equivariance in deep learning follow a basic equivariant assumption to preserve the structure of the original input, i.e., the changes in the original input will be reflected in the output in an explicit form \cite{cohen2016group}. More formally, let $g$ denote the representation extractor from the input space to a latent space, and $t \in \mathcal{T}$ denote any transformation in the input space. The equivariance property is defined as follows:
\begin{equation}
\label{eq:equivariance}
\forall t \in \mathcal{T},\exists t', \quad g(t(x)) = t'(g(x)),
\end{equation}
where $t'$ is an underlying transformation in the latent space. 
In other words, equivariance ensures that encoding the input with $g$ and then applying transformation $t'$ should give the same result as applying transformation $t$ to the original input before encoding it with $g$. By enforcing this constraint, the neural network can generalize better by maintaining the structure between the input and the output in the face of unknown transformations \cite{cohen2016group,geirhos2020shortcut}.

\subsection{Connection between Mixup and Equivariance}
\label{sec:connection}
To build the connection between the mixup and equivariance, we first generalize previous mixup-based methods within a unified framework, and then show that previous mixup-based methods share the same key idea as equivariance in deep learning. 

Let $M_\lambda^\mathcal{X}$ denote the predefined mixup transformation function in the input space with a random parameter $\lambda$ sampled from a beta distribution. Previous mixup variants have utilized $M_\lambda^\mathcal{X}$ to create a mixed input as follows  
\begin{equation}
\label{eq:generalmix}
\widetilde{x}_{i,j}=M^\mathcal{X}_\lambda(x_i,x_j),
\end{equation}
where $\widetilde{x}_{i,j}$ can be seen as a mixture of $x_i$ and $x_j$ in proportions of $\lambda$ and $1-\lambda$.
The way how the inputs are mixed in vanilla mixup (linear combination in Eq.~\ref{eq:mixup}) and CutMix (cutting and pasting in Eq.~\ref{eq:cutmix}) can be seen as special cases of this general mixup equation (Eq.~\ref{eq:generalmix}).
Now we need to set a label for the created mixed input $\widetilde{x}_{i,j}$ to supervise the neural network, so that it can give reasonable predictions when inputting mixed samples $\widetilde{x}_{i,j}$. 
To achieve this goal, as mentioned in the introduction, previous mixup variants usually share the \emph{label-equivariance assumption} to introduce the prior knowledge, i.e., 
\begin{adjustwidth}{0.5em}{0.5em}
\emph{a proportional mixup of the input data results in the corresponding labels being mixed in the same proportion.}
\end{adjustwidth}
Formally, as shown in Eq.~\ref{eq:mixup} and \ref{eq:cutmix}, previous mixup-based methods usually implicitly assume that the label $\widetilde{y}_{i,j}$ of mixed input $\widetilde{x}_{i,j}$ satisfies the following equation,
\begin{equation}
\footnotesize
\label{eq:generalmix_label}
\textcolor{red}{f^\star(M^\mathcal{X}_\lambda(x_i, x_j))} = \widetilde{y}_{i,j} \triangleq M_\lambda^\mathcal{Y}(y_i,y_j) =\textcolor{red}{M_\lambda^\mathcal{Y}(f^\star(x_i),f^\star(x_j))},
\end{equation}
where $f^\star$ is an ideal mapping function from $\mathcal{X}$ to $\mathcal{Y}$, and $M_\lambda^\mathcal{Y}$ is the mixup transformation function defined in the label space according to the prior knowledge. As shown in Eq.~\ref{eq:mixup} and Eq.~\ref{eq:cutmix}, the previous mixup-based methods typically construct $M_\lambda^\mathcal{Y}$ using linear interpolation between different labels. Based on the above artificially defined transformations $M_\lambda^\mathcal{X}$ and $M_\lambda^\mathcal{Y}$ in the input and the label space respectively, different mixup variants construct mixed training samples for regularizing the neural network.

\textbf{Similarities between mixup and equivariance.} By comparing the red parts of Eq.~\ref{eq:generalmix_label} with Eq.~\ref{eq:equivariance}, we can observe that the fundamental idea behind mixup for constructing mixed samples is similar to equivariance in deep learning. Specifically, previous mixup-based methods encourage the mixed label to vary in a well-defined manner with the transformation applied to the mixed input, i.e., transforming a pair of inputs $x_i$ and $x_j$ using the mixup transformation function $M_\lambda^\mathcal{X}$ and then passing it through the classifier mapping should yield the same result as first mapping $x_i$ and $x_j$ using $f^\star$ and then performing the label mixup transformation $M_\lambda^\mathcal{Y}$. Therefore, both mixup and equivariance aim to train the neural network to preserve the underlying structure between the input and output data, i.e., transformations on the input can be preserved in some way on the output.


\textbf{Differences between mixup and equivariant neural network.} Although mixup is closely related to equivariance, there are still the following differences in detail. \textit{Different goals.} Equivariant neural networks usually focus on designing an equivariant neural network to satisfy Eq.~\ref{eq:equivariance} strictly, while mixup implicitly constraint the neural network to satisfy Eq.~\ref{eq:generalmix_label} by creating more training samples. \textit{Different transformation.} Different from equivariant neural networks, mixup employs the transformation function $M_{\lambda}^{\mathcal{X}}$ and $M_{\lambda}^{\mathcal{Y}}$ predefined on the sample pairs to enhance the equivariance of the model.
\textit{Different encoder.} The mapping function in mixup is a deep classification network, while in equivariant neural networks, it usually is a representation extractor $g$,  which shows that the previous mixup only focuses on equivariance in the label space and ignores other semantic information.

\begin{figure}[!htbp]
\centering
\includegraphics[width=1.\linewidth,height=0.34\linewidth]{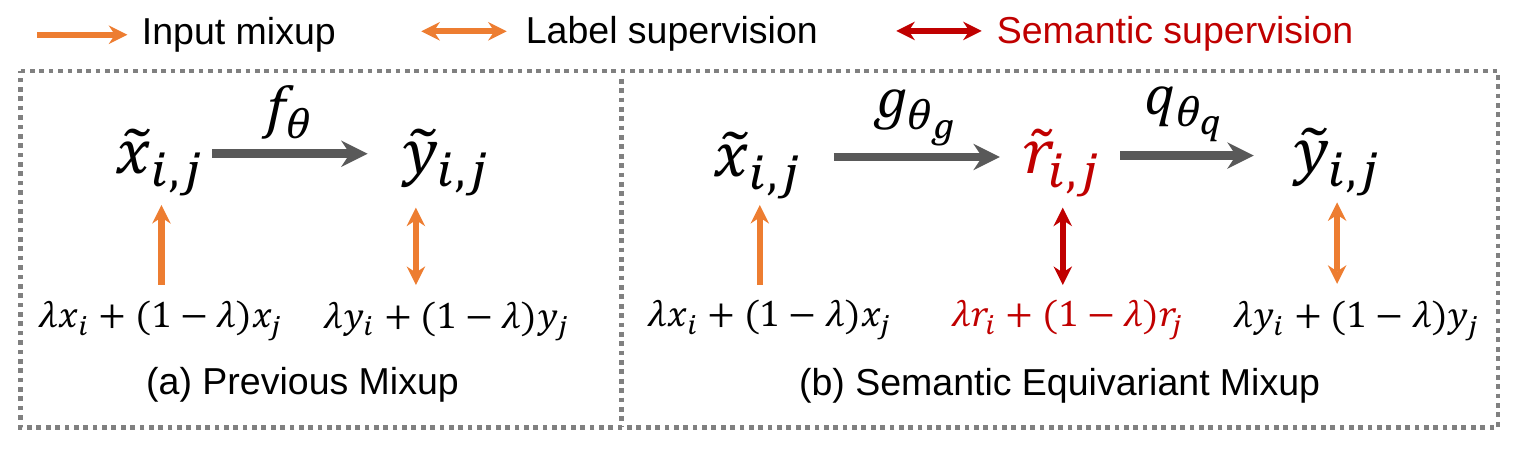}
\caption{\label{fig:framework2} The overall framework of previous mixup and the proposed method. As shown in Fig.~\ref{fig:framework2}(a), the previous mixup \cite{zhang2018mixup} creates mixed samples to train the deep neural network $f_\theta$ only with the label information. In contrast, the proposed \textsc{sem} further exploits the semantic information in the mixed samples with semantic supervision (red part in Fig.~\ref{fig:framework2}(b)).}
\end{figure}

\subsection{Semantic equivariant mixup}
\label{sec:method}
Inspired by equivariance in deep learning, we find that previous mixup variants over-focus on the label information and cannot fully exploit the semantic information in the mixed samples. However, recent studies have pointed out that over-focusing on the label information in the input may lead to the simplicity bias \cite{shah2020pitfalls,pezeshki2021gradient} and taking shortcuts \cite{geirhos2020shortcut} in the deep neural network, i.e., the model relies too much on simple features associated with labels in the input while ignoring other semantic information. As a result, when faced with samples with distributional shifts during test, the performance of the model may decrease significantly due to the removal of superficial correlations between the input and label. 

To further improve previous mixup variants, we aim to exploit the semantic information in mixed samples by generalizing the previous label-equivariance assumption in the representation space. Specifically, we first propose the semantic-equivariance assumption, i.e.,
\begin{adjustwidth}{0.5em}{0.5em}
\emph{a proportional mixup of the input results in a combination of the representations in the same proportion.}
\end{adjustwidth}
Formally, we consider that the ideal classifier can be decomposed as a representation extractor $g^\star$ and a linear classifier $q^\star$, i.e., $f^\star=g^\star\cdot q^\star$.
Then the proposed generalized semantic-equivariance assumption can be formalized as follows:
\begin{equation}
\footnotesize
\label{eq:semantic_e}
\widetilde{r}_{i,j} =  g^\star(M_{\lambda}^\mathcal{X}(x_i,x_j))=M_\lambda^\mathcal{R}(g^\star(x_i),g^\star(x_j)) = M_\lambda^\mathcal{R}(r_i,r_j),
\end{equation}
where $M_\lambda^\mathcal{R}$ is the predefined transformation function in the representation space and we can define it as a linear interpolation between $r_i$ and $r_j$,
\begin{equation}
M_\lambda^\mathcal{R}(r_i,r_j)=\lambda r_i + (1-\lambda) r_j.
\end{equation}
Intuitively, we assume that the representation of the mixed sample should be consistent with the representations' mixture of the original samples. Different from previous mixups which only assume the equivariance of label information, we consider the semantic information of mixed samples to be equivariant as well.

Now we consider how to leverage the semantic equivariance assumption to cooperate with the previous label-equivariance assumption to promote the training of the deep neural network $f_\theta$. Let $f_\theta$ be decomposed into a feature extractor $g_{\theta_g}$ and a linear classifier $q_{\theta_q}$ with parameter $\theta_g$ and $\theta_q$ respectively. Similar to the previous mixup loss, we can regularize the neural network $f_\theta$ with the following loss,
\begin{equation}
\footnotesize
\label{eq:semloss}
\mathbb{E}_\mathcal{\widetilde{D}}\{\underbrace{\ell[f_{\theta}(\widetilde{x}_{i,j}),\widetilde{y}_{i,j}]}_{\mbox{label\ supervision}}+\gamma\underbrace{\|g_{\theta_g}(\widetilde{x}_{i,j})-M_{\lambda}^{\mathcal{R}}[g_{\theta_g}(x_i),g_{\theta_g}(x_j)]\|_2}_{\mbox{semantic\ supervision}}\},
\end{equation}
where $\mathcal{\widetilde{D}}$ is the dataset composed of mixed samples and $\gamma$ is a hyperparameter that balances different losses. The first term in Eq.~\ref{eq:semloss} can be regarded as the loss of the original mixup, and the second term is the regularization term proposed in this paper to utilize the semantic information in the mixed samples. In fact, it is intractable to design networks that strictly satisfy semantic equivariance, so similar to previous works \cite{devillers2023equimod,dangovski2022equivariant}, Eq.~\ref{eq:semloss} can promote semantic equivariance of the neural network $f_\theta$ with a mild regularization. The overall framework and the training pseudocode of \textsc{sem} is shown in the Fig.~\ref{fig:framework2} and Alg.~\ref{alg:semix} respectively.

\begin{algorithm}[!tb]
\caption{The training pseudocode of \textsc{sem}.\label{algo}\label{alg:semix}}
\textbf{Input}:Training dataset $\mathcal{D}$, hyperparameter $\alpha$ of the beta distribution, hyperparameter $\gamma$; \\
\textbf{Output}: Neural network $f_{\theta}$; 
\begin{algorithmic}[1] 
    \FOR{Each iteration}
    \STATE Obtain training samples $(x_{i}, y_{i})$, $(x_{j}, y_{j})$;
    \STATE Sample $\lambda \sim Beta(\alpha, \alpha)$ and create mixed sample $(\widetilde{x}_{i,j}, \widetilde{y}_{i,j})$ with previous mixup methods;
    \STATE Obtain the representation $\widetilde{r}_{i,j}$, $r_i$ and $r_j$ with feature extractor $g_{\theta_g}$;
    \STATE Obtain the prediction results of $\widetilde{x}_{i,j}$ with $f_{\theta}$;
    \STATE Obtain the loss of the model with Eq.~\ref{eq:semloss};
    \STATE Update model parameters $\theta$ to minimize loss with an optimization algorithm.
    \ENDFOR
\end{algorithmic}
\end{algorithm}

\textbf{Rethink why the proposed method works}. The proposed method aims to exploit semantic information in mixed samples to improve the previous mixup, which has several advantages: (1) More supervision information: compared to previous mixup, \textsc{sem} provides more comprehensive supervision at the semantic level. (2) Stronger regularization: by leveraging semantic information in mixed samples, the model is better regularized to prevent overfitting. (3) Better representation: \textsc{sem} forces the learned representations to retain more semantic information rather than just focusing on input-label mapping, allowing the model to understand more complex data. (4) Less shortcut learning: semantic information compels the model to learn essential mapping relations instead of relying on labels only, avoiding shortcut solutions.

%% file: 4_experiments.tex
\begin{table}[!bp]
\footnotesize
  \centering
  \caption{\textbf{Q1 (Effectiveness).} Comparison results with other methods on multiple datasets and backbone networks. Combining the proposed \textsc{sem} with the previous mixup variants can significantly improve the model performance.}
    \begin{tabular}{p{4.3em}p{1.25em}<{\centering}p{1.25em}<{\centering}p{1.25em}<{\centering}p{1.25em}<{\centering}p{1.25em}<{\centering}p{1.25em}<{\centering}}
    \toprule
    Datasets / & \multicolumn{3}{c}{PreActResNet18} & \multicolumn{3}{c}{WideResNet-28-10} \\
    ACC(\%) & \multicolumn{1}{c}{\footnotesize{C10}} & \multicolumn{1}{c}{\footnotesize{C100}} & \multicolumn{1}{c}{\scriptsize{TImage}} & \multicolumn{1}{c}{\footnotesize{C10}} & \multicolumn{1}{c}{\footnotesize{C100}} & \multicolumn{1}{c}{\scriptsize{TImage}} \\
    \midrule
    ERM   & 94.62 & 74.97 & 59.41 & 96.40 & 79.06 & 65.59 \\
    \midrule
    mixup & 95.77 & 77.44 & 60.39 & 96.99 & 82.33 & 67.30 \\
    mixup+\textsc{sem} & 96.45 & 79.31 & 63.99 & 97.36 & 83.13 & 69.62 \\
    \scriptsize{Improvement} & \textcolor{blue}{+0.68} & \textcolor{blue}{+1.84} & \textcolor{blue}{+3.60} & \textcolor{blue}{+0.37} & \textcolor{blue}{+0.80} & \textcolor{blue}{+2.32} \\
    \midrule
    mixup+ES & 95.85 & 79.49 & 61.10 & 97.18 & 83.24 & 69.18 \\
    \scriptsize{mixup+ES+\textsc{sem}} & 96.14 & 80.67 & 65.11 & 97.42 & 83.65 & 70.61 \\
    \scriptsize{Improvement} & \textcolor{blue}{+0.29} & \textcolor{blue}{+1.18} & \textcolor{blue}{+4.01} & \textcolor{blue}{+0.24} & \textcolor{blue}{+0.41} & \textcolor{blue}{+1.43} \\
    \midrule
    CutMix & 95.94 & 79.57 & 65.20 & 97.04 & 82.68 & 70.18 \\
    CutMix+\textsc{sem} & 96.45 & 80.18 & 65.90 & 97.28 & 83.15 & 71.26 \\
    \scriptsize{Improvement} & \textcolor{blue}{+0.51} & \textcolor{blue}{+0.61} & \textcolor{blue}{+0.70} & \textcolor{blue}{+0.24} & \textcolor{blue}{+0.47} & \textcolor{blue}{+1.08} \\
    \midrule
    CutMix+ES & 96.32 & 79.70 & 65.59 & 96.99 & 83.05 & 70.25 \\
    \scriptsize{CutMix+ES+\textsc{sem}} & 96.34 & 80.30 & 66.66 & 97.14 & 83.44 & 71.46 \\
    \scriptsize{Improvement} & \textcolor{blue}{+0.02} & \textcolor{blue}{+0.60} & \textcolor{blue}{+1.07} & \textcolor{blue}{+0.15} & \textcolor{blue}{+0.39} & \textcolor{blue}{+1.21} \\
    \bottomrule
    \end{tabular}%
  \label{tab:imagecls}%
\end{table}%

\begin{table*}[!tbp]
\small
  \centering
  \caption{\textbf{Q2 (Generalization).} Comparison results on multiple datasets data with corruptions and natural covariate shifts.}
    \begin{tabular}{l|cccc|cccc}
    \toprule
    Datasets/ & \multicolumn{4}{c|}{PreActResNet18}    & \multicolumn{4}{c}{WideResNet-28-10} \\
    ACC(\%) & C10-C & C10.1 & C10.2 & C100-C & C10-C & C10.1 & C10.2 & C100-C \\
    \midrule
    ERM   & 75.08 & 86.98 & 81.95 & 49.68 & 78.01 & 90.30 & 84.70 & 52.17 \\
    \midrule
    Mixup & 78.80 & 89.41 & 84.84 & 55.03 & 82.59 & 92.03 & 87.65 & 59.93 \\
    Mixup+\textsc{sem} & 79.34 & 90.20 & 85.70 & 57.12 & 82.70 & 92.13 & 87.75 & 61.01 \\
    Improvement & \textcolor{blue}{+0.54} & \textcolor{blue}{+0.79} & \textcolor{blue}{+0.86} & \textcolor{blue}{+2.09} & \textcolor{blue}{+0.11} & \textcolor{blue}{+0.10} & \textcolor{blue}{+0.10} & \textcolor{blue}{+1.08} \\
    \midrule
    Mixup+ES & 80.02 & 90.10 & 85.20 & 55.83 & 81.33 & 92.28 & 87.30 & 60.49 \\
    Mixup+ES+\textsc{sem} & 81.38 & 90.45 & 86.21 & 57.21 & 83.32 & 92.78 & 87.86 & 62.08 \\
    Improvement & \textcolor{blue}{+1.36} & \textcolor{blue}{+0.35} & \textcolor{blue}{+1.01} & \textcolor{blue}{+1.38} & \textcolor{blue}{+1.99} & \textcolor{blue}{+0.50} & \textcolor{blue}{+0.56} & \textcolor{blue}{+1.59} \\
    \midrule
    CutMix & 73.34 & 90.40 & 85.20 & 49.63 & 75.14 & 91.14 & 87.35 & 52.25 \\
    CutMix+\textsc{sem} & 74.48 & 91.14 & 85.10 & 50.08 & 75.64 & 92.49 & 87.70 & 52.81 \\
    Improvement & \textcolor{blue}{+1.14} & \textcolor{blue}{+0.74} & \textcolor{brown}{-0.10} & \textcolor{blue}{+0.45} & \textcolor{blue}{+0.50} & \textcolor{blue}{+1.35} & \textcolor{blue}{+0.35} & \textcolor{blue}{+0.56} \\
    \midrule
    CutMix+ES & 73.89 & 90.80 & 84.35 & 49.91 & 75.26 & 91.09 & 87.15 & 51.63 \\
    CutMix+ES+\textsc{sem} & 75.12 & 90.65 & 85.30 & 50.44 & 75.67 & 92.13 & 87.75 & 52.57 \\
    Improvement & \textcolor{blue}{+1.23} & \textcolor{brown}{-0.15} & \textcolor{blue}{+0.95} & \textcolor{blue}{+0.53} & \textcolor{blue}{+0.41} & \textcolor{blue}{+1.04} & \textcolor{blue}{+0.60} & \textcolor{blue}{+0.94} \\
    \bottomrule
    \end{tabular}%
  \label{tab:cifarcorruption}%
\end{table*}%

In this section, we comprehensively evaluate \textsc{sem} by investigating its improvements over the following aspects:
\begin{itemize}
    \item \textbf{Q1 (Effectiveness):} Does the proposed method outperform the classical mixup methods on image classification task?
    \item \textbf{Q2 (Generalization):} Can the proposed method improve the model's generalization to data with distribution shifts?
    \item \textbf{Q3 (OOD detection):} Can the proposed method improve the performance of the model for out-of-distribution (OOD) detection?
    \item \textbf{Q4 (Qualitative analysis):} How does the representation of the obtained mixed samples differ from previous mixup variants?
\end{itemize}

\subsection{Experimental Setup}
In this section, we present the details of our experimental setup, including comparison methods, experimental datasets, neural network architectures, and hyperparameter settings.

\textbf{Comparison methods.} To verify the effectiveness of the proposed method, we perform comparative experiments from the following three perspectives. 
(1) Comparison with empirical risk minimization (ERM). We compare the proposed method with empirical risk minimization to show that using mixed samples can improve the performance of the model. (2) Comparison with vanilla mixup \cite{zhang2018mixup} and CutMix \cite{yun2019cutmix}. Without loss of generality, we employ vanilla mixup and CutMix to create mixed samples and apply our method to them to verify that the proposed method can improve its performance. Note that the proposed method is a general framework that can cooperate with most mixup variants and not limited to vanilla mixup and CutMix. (3) We combine the proposed method with mixup's early stop (ES) strategy \cite{liu2023over,zou2023benefits} to verify the effect of the proposed method with other mixup improvement methods, where early stop refers to not using mixed samples in the final stage of training.

\textbf{Datasets.} We perform experiments on widely used datasets, including CIFAR10 (C10), CIFAR100 (C100) \cite{krizhevsky2009learning} and Tiny-ImageNet (TImage) \cite{chrabaszcz2017downsampled}. To verify the generalization of the proposed method against distribution shifts, we evaluate the model on corruption datasets and datasets with covariate shifts. Specifically, we evaluate the model on CIFAR10-C and CIFAR100-C datasets \cite{hendrycks2019benchmarking}, which are made by adding 15 kinds of corruptions to CIFAR10 and CIFAR100 with different degree of intensity. Besides, to better test generalization of the model, we also use the CIFAR10.1 \cite{recht2018cifar} and CIFAR10.2 \cite{lu2020harder} datasets as covariate shifts datasets. For OOD detection, we choose SVHN \cite{netzer2011reading} and LSUN-Crop \cite{yu2015lsun} as OOD datasets and set CIFAR-10 and CIFAR-100 as in-distribution datasets.

\textbf{Network Architectures.} We conduct experiments on two commonly used backbone networks including WideResNet-28-10 \cite{zagoruyko2016wide} and PreActResNet18 \cite{he2016identity}. For all backbone networks we train for 200 epochs on all datasets. 

\textbf{Hyperparameter Settings.} We train the deep neural network with the SGD optimizer with learning rate decay and weight decay. We choose the hyperparameter $\alpha$ of the beta distribution from $\{0.2,1,2,5\}$ and the $\gamma$ from $\{0.1, 0.2, ... ,1\}$. 

\subsection{Experimental results}
\textbf{Q1 Effectiveness.} We conduct comparison experiments and report the classification accuracy on CIFAR10, CIFAR100, and TinyImageNet datasets to investigate the effectiveness of \textsc{sem}. The experimental results are shown in Tab.~\ref{tab:imagecls} and we have the following observations. (1) Compared with ERM, the methods using mixed samples are all able to significantly improve the performance of the model. For example, when using PreActResNet18 as the backbone model, mixup with \textsc{sem} improves ERM by 1.83\%, 4.93\% and 4.58\% on CIFAR10, CIFAR100, and Tiny ImageNet, respectively. (2) Benefiting from exploiting semantic information in mixed samples, combining variants of the previous mixup methods with \textsc{sem} can significantly improve the performance of the model on all datasets and architectures. For example, when using PreActResNet18 as the backbone, our method gains 0.68\% on CIFAR10, 1.87\% on CIFAR100, and 3.6\% on Tiny-ImageNet over Mixup.

\textbf{Q2 Generalization.} To demonstrate generalization of our method, we first conduct experiments on corrupted datasets including  CIFAR10-C (C10-C) and CIFAR100-C (C100-C). Following the previous work \cite{hendrycks2019benchmarking}, we report the average accuracy over all types of corruptions and all degrees of intensities. Besides, we also evaluate the model on the datasets with natural covariate shifts, i.e., CIFAR10.1 (C10.1) and CIFAR10.2 (C10.2). The full experimental results are shown in Tab.~\ref{tab:cifarcorruption}. From the experimental results, we have the following observations. (1) The proposed method can improve generalization performance of previous mixup methods on corrupted datasets. For instance, for C100-C with PreActResNet18, our method gains 2.09\% over Mixup and 0.45\% over Cutmix.  (2) The proposed method can further improve the generalization of the model to data with natural covariate shift by exploiting the semantic information in the mixed samples. For example, when using PreActResNet18, the proposed method improves Mixup by 0.79\% and 0.86\% on C10.1 and C10.2 datasets, respectively.

\begin{table*}[!htbp]
\small
  \centering
  \caption{\textbf{Q3 (OOD detection).} Experimental results for out-of-distribution samples detection. }
    \begin{tabular}{l|cc|cc|cc|cc}
    \toprule
          & \multicolumn{4}{c|}{PreActResNet18}    & \multicolumn{4}{c}{WideResNet-28-10} \\
    ID & \multicolumn{2}{c}{C10} & \multicolumn{2}{c|}{C100} & \multicolumn{2}{c}{C10} & \multicolumn{2}{c}{C100} \\
    \midrule
    OOD & \multicolumn{1}{l}{SVHN} & \multicolumn{1}{l|}{LSUN} & \multicolumn{1}{l}{SVHN} & \multicolumn{1}{l|}{LSUN} & \multicolumn{1}{l}{SVHN} & \multicolumn{1}{l|}{LSUN} & \multicolumn{1}{l}{SVHN} & \multicolumn{1}{l}{LSUN} \\
    \midrule
    ERM   & 84.88 & 94.66 & 75.87 & 80.04 & 89.56 & 95.73 & 79.08 & 82.59 \\
    \midrule
    mixup & 89.31 & 92.69 & 74.48 & 79.32 & 92.34 & 92.69 & 77.51 & 81.96 \\
    mixup+\textsc{sem} & 92.54 & 97.10 & 75.29 & 82.28 & 93.42 & 95.77 & 86.34 & 85.56 \\
    Improvement & \textcolor{blue}{+3.23} & \textcolor{blue}{+4.41} & \textcolor{blue}{+0.81} & \textcolor{blue}{+2.96} & \textcolor{blue}{+1.08} & \textcolor{blue}{+3.08} & \textcolor{blue}{+8.83} & \textcolor{blue}{+3.60} \\
    \midrule
    mixup+ES & 93.03 & 96.41 & 78.46 & 83.18 & 97.26 & 95.22 & 80.70  & 87.14 \\
    mixup+ES+\textsc{sem} & 94.97 & 96.34 & 80.29 & 87.34 & 96.06 & 96.33 & 84.07 & 87.96 \\
    Improvement & \textcolor{blue}{+1.94} & \textcolor{brown}{-0.07} & \textcolor{blue}{+1.83} & \textcolor{blue}{+4.16} & \textcolor{brown}{-1.20} & \textcolor{blue}{+1.11} & \textcolor{blue}{+3.37} & \textcolor{blue}{+0.82} \\
    \midrule
    CutMix & 90.61 & 98.10 & 76.41 & 89.21 & 90.80  & 98.09 & 71.32 & 82.41 \\
    CutMix+\textsc{sem} & 95.93 & 97.94 & 82.04 & 90.55 & 93.92 & 96.01 & 86.75 & 86.81 \\
    Improvement & \textcolor{blue}{+5.32} & \textcolor{brown}{-0.16} & \textcolor{blue}{+5.63} & \textcolor{blue}{+1.34} & \textcolor{blue}{+3.12} & \textcolor{brown}{-2.08} & \textcolor{blue}{+15.43} & \textcolor{blue}{+4.40} \\
    \midrule
    CutMix+ES & 95.21 & 96.20  & 84.88 & 85.47 & 91.39 & 93.15 & 81.43 & 84.99 \\
    CutMix+ES+\textsc{sem} & 94.02 & 96.35 & 80.75 & 85.59 & 92.72 & 95.31 & 81.20 & 84.75 \\
    Improvement & \textcolor{brown}{-1.19} & \textcolor{blue}{+0.15} & \textcolor{brown}{-4.13} & \textcolor{blue}{+0.12} & \textcolor{blue}{+1.33} & \textcolor{blue}{+2.16} & \textcolor{brown}{-0.23} & \textcolor{brown}{-0.24} \\
    \bottomrule
    \end{tabular}%
  \label{tab:OOD}%
\end{table*}%

\begin{figure}[!tbp]
    \hspace{-10pt}
    \begin{minipage}[t]{0.5\linewidth}
        \centering
        \includegraphics[width=\textwidth]{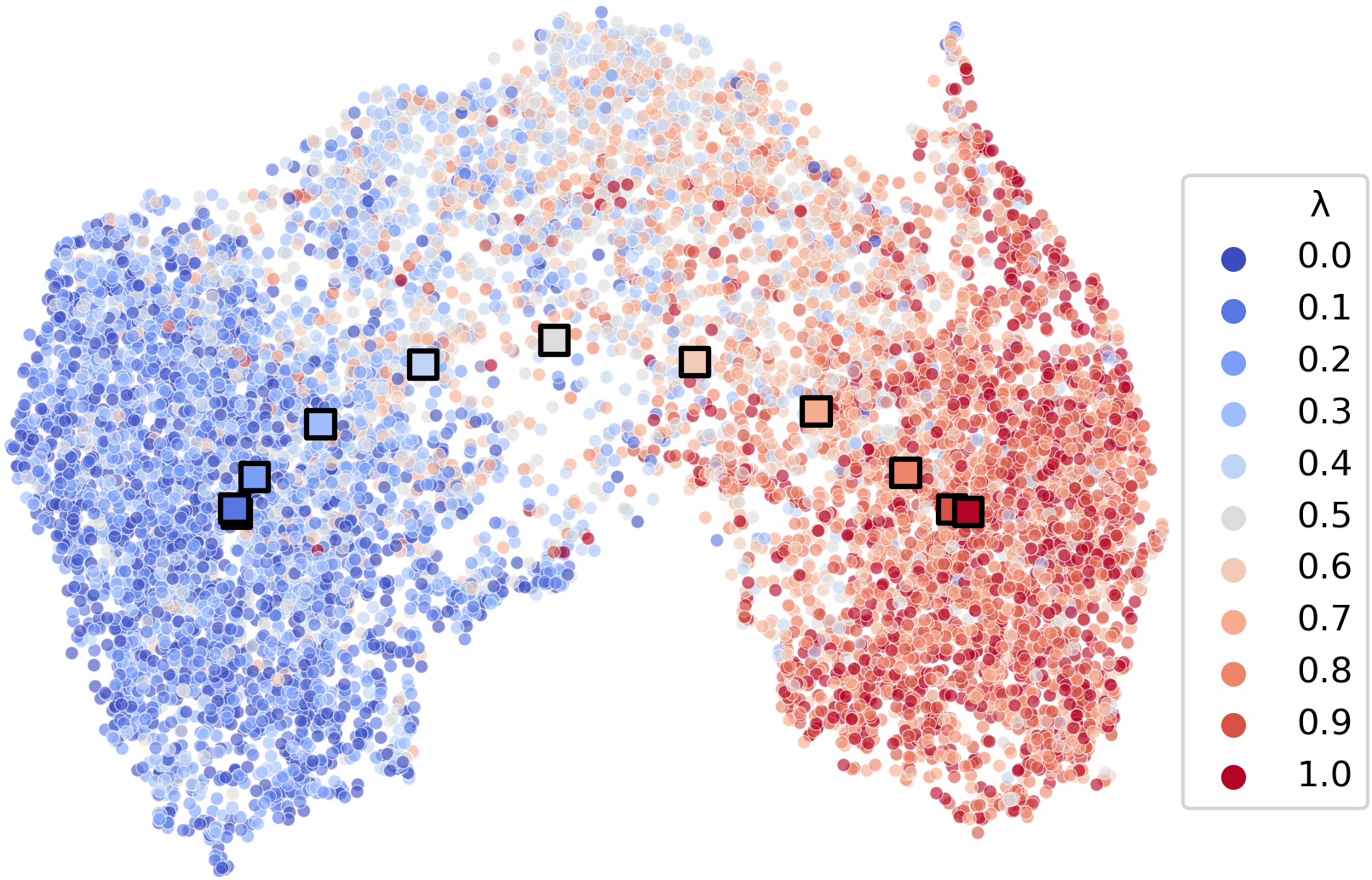}
        \centerline{(a) mixup}
    \end{minipage}%
    \hspace{5pt}
    \begin{minipage}[t]{0.5\linewidth}
        \centering
        \includegraphics[width=\textwidth]{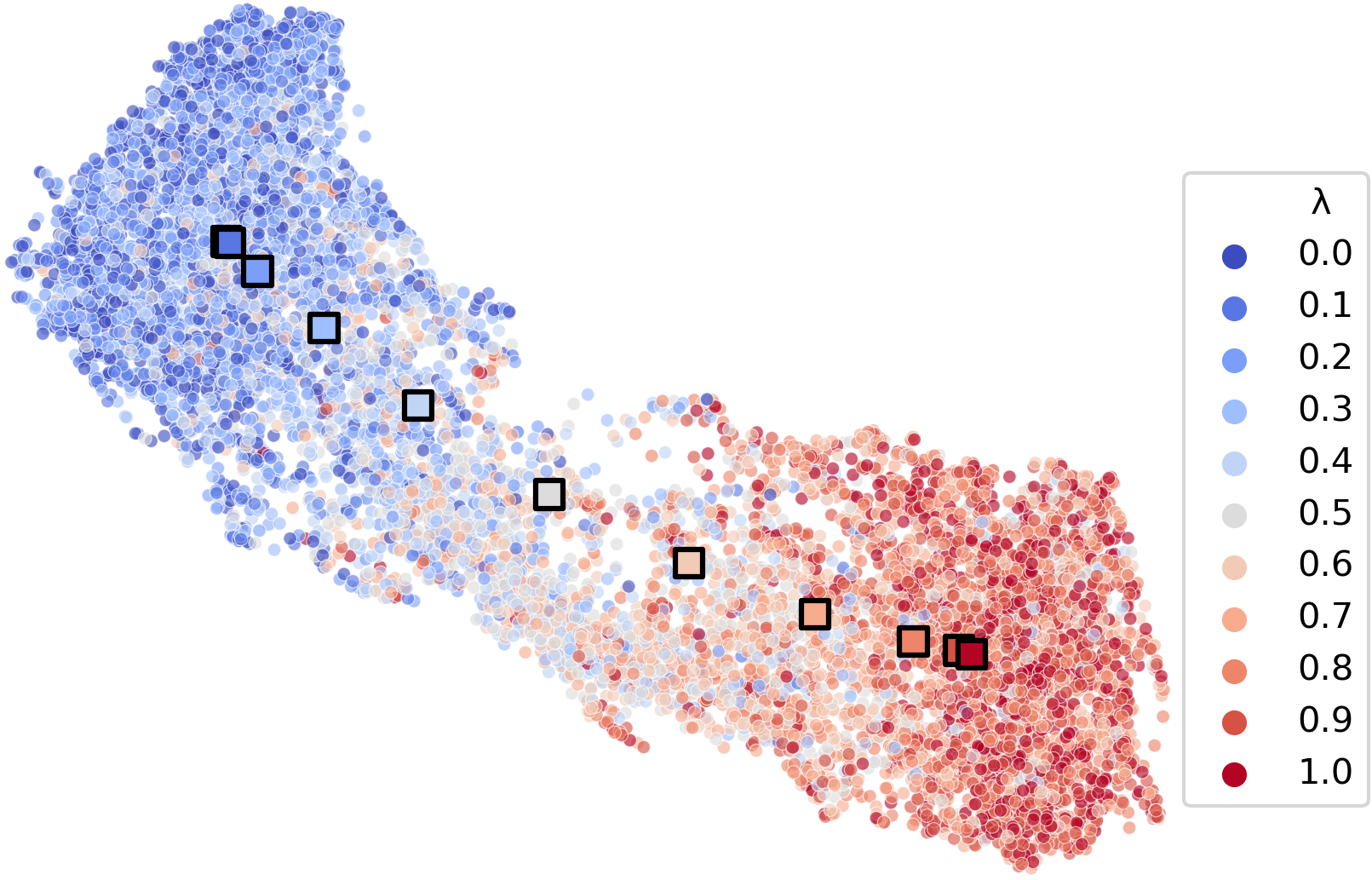}
        \centerline{(b) mixup+\textsc{sem}}
    \end{minipage}
    \caption{\textbf{Q4 (Qualitative analysis).} Visualization of the representation of mixed samples with different mixing  proportions $\lambda$. It can be found that the representation obtained by mixup+\textsc{sem} better satisfies the semantic equivariance assumption in Eq.~\ref{eq:semantic_e}.} 
    \label{fig:umap}
\end{figure}
\textbf{Q3 OOD detection.} To further evaluate the robustness of the proposed method to out-of-distribution data, we conduct out-of-distribution detection experiments. Specifically, we use CIFAR10 and CIFAR100 as in-distribution data, and SVHN and LSUN as out-of-distribution data, respectively. We directly use the maximum softmax probability of the classifier as the score to distinguish ID data from OOD data, and use AUROC as the metric for OOD detection.  The experimental results are reported in Tab.~\ref{tab:OOD} and from the experimental results, we can draw the following conclusion. The proposed method can enhance the detection ability of the model for out-of-distribution samples in most cases. For example, compared with vanilla mixup, the mixup with semantic equivariance regularization can significantly improve the performance of OOD detection, and it has achieved an average performance improvement of 3.5\% in all ID-OOD detection tasks. The fundamental reason is that the proposed method is able to enhance the information in the representation by exploiting the semantic information in the mixed samples, thus preserving the structure between input and output to facilitate the classifier to distinguish in-distribution data from OOD data. 

\textbf{Q4 Qualitative analysis.}
To qualitatively analyze the effect of \textsc{sem}, we visualize the representations of the mixed samples to compare the \textsc{sem} 's representations with the previous mixup representations. Specifically, for a better presentation, we randomly select two classes from CIFAR10, and mix the samples of the two classes in different proportions (i.e, $\lambda=[0,0.1,\cdots,1]$) to obtain mixed samples with Eq.~\ref{eq:mixup}. Then we use the feature extractor obtained by vanilla mixup training and training with \textsc{sem} regularization to obtain the representation of the mixed sample. Finally we visualize the obtained representations with UMAP \cite{mcinnes2018umap} and the experimental results are shown in Fig.~\ref{fig:umap}. From the experimental results we can draw the following conclusions. (1) The representations of mixed samples obtained by vanilla mixup training are semantically inequivalent, i.e., as shown in Fig.~\ref{fig:umap}(a), the representations of mixed samples are different from the mixture of original sample representations. (2) When the vanilla mixup is cooperated with the proposed \textsc{sem}, the obtained mixed sample representation has better semantic equivariance. Specifically, as shown in Fig.~\ref{fig:umap}(b), the representation of the mixed sample is closer to the combination of the representation of the original samples.
